\documentclass[11pt]{article}

\usepackage[final]{acl}
\usepackage[dvipsnames]{xcolor}

\usepackage{times}
\usepackage{xurl}
\usepackage{latexsym}
\usepackage{float} 
\usepackage{booktabs}
\usepackage[T1]{fontenc}
\usepackage{xcolor}

\usepackage[utf8]{inputenc}

\usepackage{microtype}

\usepackage{inconsolata}

\usepackage{graphicx}

 \usepackage{tikz}
\usetikzlibrary{arrows.meta,positioning,fit,shapes}

\title{Defining Cultural Capabilities for AI Evaluation: A Taxonomy Grounded in Intercultural Communication Theory}

\author{
  Isar Nejadgholi\textsuperscript{\rm 1}, Masoud Kianpour\textsuperscript{\rm 2}\\
  \bf{Krishnapriya Vishnubhotla}\textsuperscript{\rm 1}, \bf{Maryam Molamohamadi}\textsuperscript{\rm 3}\\ \\
  \textsuperscript{\rm 1}National Research Council, Canada \quad
  \textsuperscript{\rm 2}Toronto Metropolitan University, Canada\\
  \textsuperscript{\rm 3}Mila, Quebec AI Institute, Canada\\
  \small\texttt{\{isar.nejadgholi, krishnapriya.vishnubhotla\}@nrc-cnrc.gc.ca}\\
  \small\texttt{masoud.kianpour@torontomu.ca, maryam.molamohammadi@mila.quebec}
}

\begin{document}

\maketitle

\begin{abstract}
 Tremendous efforts have been put into evaluating the inclusivity and effectiveness of AI systems across cultures. However, the cultural capabilities considered in much of the literature remain vaguely defined, are referred to using interchangeable terminology, and are typically limited to recalling accurate information about various demographics, regions, and nationalities. To address this construct ambiguity, we draw from Intercultural Communication scholarship and propose a three-level taxonomy of AI-relevant cultural capabilities: \textbf{Cultural Awareness} answers \emph{``Does the model know?''}, \textbf{Cultural Sensitivity} answers \emph{``How does it frame its knowledge?''}, and \textbf{Cultural Competence} answers \emph{``Can it adapt as the interaction evolves?''}. Beyond conceptual clarification, we position this taxonomy as a practical tool for improving the validity and interpretability of AI evaluation in real-world, multicultural settings. Without such construct clarity, evaluation results risk overstating model capabilities and may lead to inappropriate deployment decisions in culturally sensitive contexts.


\end{abstract}

\section{Introduction}
\label{sec:intro}

AI-mediated communication is increasingly impacting language and social relationships \cite{hohenstein2023artificial}. In a variety of tasks, such as translation \cite{naveen2024overview}, dialogue \cite{abe2025classification}, and decision-making \cite{kaggwa2024ai}, AI is mediating conversations among users from every corner of the globe, across cultural boundaries. Generative AI in particular has been shown to act as a ``social actor,'' capable of eliciting emotional and cognitive responses that reshape human communication patterns. The research community, however, is coming to an understanding that the impact of generative AI on human communication is extremely nuanced. On the one hand, research shows that AI can enhance cross-cultural dialogue by providing multimodal, emotionally resonant communication tools that reduce anxiety and facilitate identity recognition \cite{yang2024impact}. On the other hand, when used at scale, AI introduces new dynamics of power and cultural visibility that risk homogenizing cultural expressions, reinforcing linguistic hierarchies, and obscuring subtle cultural meanings \cite{busch2024ai}. Crucially, these models are primarily trained on English- and Western-centric data, which limits their abilities in handling intercultural communications and risks misunderstandings that escalate into real social and ethical harms \cite{naous2025origin}.

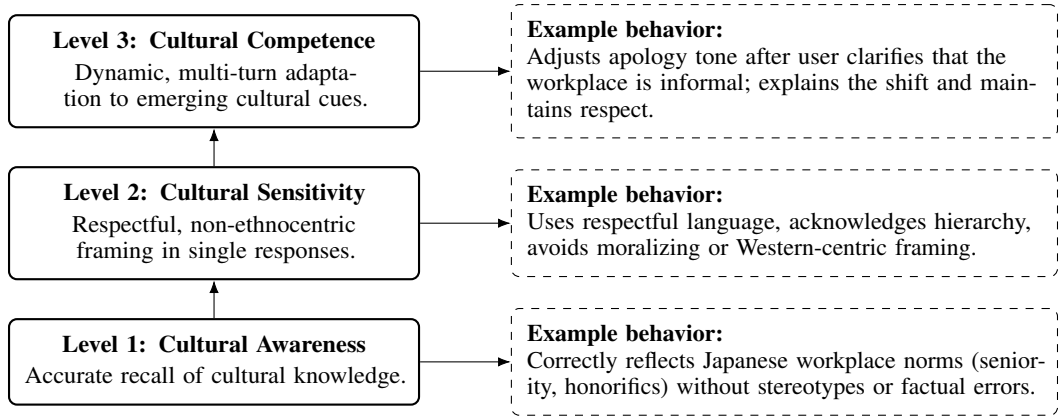
\begin{figure*}[t]
    \centering
    \begin{tikzpicture}[
        node distance=0.5cm and 1.2cm,
        box/.style={
            rectangle,
            rounded corners=3pt,
            draw,
            thick,
            align=center,
            inner sep=6pt,
            text width=5cm,
            font=\small
        },
        exbox/.style={
            rectangle,
            rounded corners=3pt,
            draw,
            dashed,
            align=left,
            inner sep=6pt,
            text width=6.8cm,
            font=\small
        }
    ]

        \node[box] (competence3) {
            \textbf{Level 3: Cultural Competence}\\[2pt]
            Dynamic, multi-turn adaptation to emerging cultural cues.
        };

        \node[box, below=of competence3] (sensitivity2) {
            \textbf{Level 2: Cultural Sensitivity}\\[2pt]
            Respectful, non-ethnocentric framing in single responses.
        };

        \node[box, below=of sensitivity2] (awareness1) {
            \textbf{Level 1: Cultural Awareness}\\[2pt]
            Accurate recall of cultural knowledge.
        };

        \node[exbox, right=of competence3] (ex3) {
            \textbf{Example behavior:}\\
            Adjusts apology tone after user clarifies that the workplace is informal; explains the shift and maintains respect.
        };

        \node[exbox, right=of sensitivity2] (ex2) {
            \textbf{Example behavior:}\\
            Uses respectful language, acknowledges hierarchy, avoids moralizing or Western-centric framing.
        };

        \node[exbox, right=of awareness1] (ex1) {
            \textbf{Example behavior:}\\
            Correctly reflects Japanese workplace norms (seniority, honorifics) without stereotypes or factual errors.
        };

        \draw[-{Latex}] (awareness1.north) -- (sensitivity2.south);
        \draw[-{Latex}] (sensitivity2.north) -- (competence3.south);

        \draw[-{Latex}] (competence3.east) -- (ex3.west);
        \draw[-{Latex}] (sensitivity2.east) -- (ex2.west);
        \draw[-{Latex}] (awareness1.east) -- (ex1.west);

    \end{tikzpicture}

    \caption{Three levels of AI-relevant cultural capabilities, defined in terms of observable system behavior, with an illustrative example aligned to each level. The example is based on the prompt \textit{``I am from Japan, and I need help apologizing to my older colleague for a mistake I made at work,''} to illustrate how progressively richer cultural capabilities shape system responses from factual grounding to respectful framing and multi-turn adaptation.  }
    \label{fig:taxonomy}
\end{figure*}

In response, a growing body of work has attempted to evaluate the ``cultural capabilities'' of AI systems \cite{pawar2025survey}. However, the constructs underlying these evaluations remain loosely defined. Terms such as cultural awareness, cultural sensitivity, and cultural competence are often used interchangeably, with inconsistent meanings across studies and even within the same work. As a result, current evaluation practices risk conflating fundamentally different capabilities. This construct ambiguity makes it unclear what is being measured and what conclusions can be drawn about model behavior in real-world settings.

In this work, we engage with the fundamental question of \textit{``What cultural capabilities need to be monitored in AI-enabled communication tools, to ensure the wide range of issues arising from English-centric models are appropriately mitigated?''}. 
Importantly, fields such as intercultural communication \cite{arasaratnam2005intercultural}, cross-cultural social psychology \cite{richter2023reviewing}, and education \cite{choompunuch2024exploring} have long emphasized that cultural capability involves multiple, distinct behaviors that enable successful interaction across cultural boundaries. These capabilities have been shown to shape outcomes in organizational, professional, and educational environments, and contribute to performance, productivity, and psychological safety \cite{lauring2011intercultural, szkudlarek2020communication, warren2020intercultural}. Yet NLP evaluations rarely incorporate these distinctions, and when viewed against the backdrop of multicultural communication research, contemporary evaluations seem under-theorized. 



A systematized construct definition of cultural capabilities can facilitate meaningful AI evaluation practices. As \citet{wallach2024evaluating} argue, valid evaluation requires moving from background concepts to systematic definitions, and only then to measurement instruments. This logic suggests that before a cultural capability can be measured, it must first be defined in terms of observable system behaviors. To assemble such a definition, we focus on the research in Intercultural Communication (ICC), where cultural capabilities are formulated as a broad range of skills such as calibrating the level of sensitivity required in a given scenario, adapting to contextual cues, and incorporating new cultural information that emerges dynamically in interaction. From this perspective, an AI system does not merely need to ``know about'' a culture or ``imitate a cultural norm''; it must be able to adjust its communicative stance in a way that respects cultural variation and is contextually appropriate. 

Moreover, the distinction between cultural capabilities is critical because behavior that is appropriate at one level may be harmful at another. For example, factual knowledge about a cultural group can support representation and understanding, but when presented without nuance or contextual variation, it may function as stereotyping \cite{fraser-etal-2021-understanding,yao2024benchmarking}. An AI system that states ``Japanese workplaces value formality'' conveys accurate information; however, presenting this as a universal rule without acknowledging regional, generational, or organizational variation risks reinforcing stereotypes. Also, this factual knowledge may not translate to appropriate behavioral/situational adaptation in user interactions. 

Specifically, we turn to three foundational models in ICC and study the traits and skills included in these models. To draw an AI-relevant taxonomy, we exclude human-specific motivational and affective traits of ICC models and retain only those dimensions that describe behavioral and interactional skills that AI systems could, in principle, exhibit. This procedure results in a three-level taxonomy of AI-relevant cultural capabilities, Cultural \textbf{Awareness, Sensitivity, and Competence}, with distinct observable behaviors. This taxonomy is summarized in Figure~\ref{fig:taxonomy} and elaborated in Section~\ref{sec:taxonomy}. Our taxonomy offers a practical framework to guide evaluation design, interpretation, and deployment decisions in multicultural settings. 
We position this work as a call for more precise, practice-oriented evaluation of cultural capabilities in AI systems. 




\section{Cultural Capability Evaluation in NLP}

Many works in NLP have investigated whether LLMs demonstrate different abilities for handling cultural variation \cite{pawar2025survey}. This line of research typically evaluates model behavior across culturally situated scenarios, norms, and communication practices. However, the conceptualization of what constitutes cultural capability varies widely across studies. We review recent NLP papers that attempt to measure cultural capability in AI and analyze how these works define and operationalize the underlying constructs. Note that we focus on the construct ambiguity of ``cultural capability'', not ``culture'' itself. While the definition of ``Culture'' has been extensively studied by \citet{zhou-etal-2025-culture} and \citet{adilazuarda-etal-2024-towards}, and was addressed through taxonomies \citep{liu-etal-2025-culturally} or foundational frameworks for cross-cultural NLP \citep{hershcovich-etal-2022-challenges}, we argue that the field has yet to converge on which \emph{cultural capabilities} are essential to assess in AI systems.

 \citet{saha-etal-2025-meta} critically examine how cultural capability in AI systems should be conceptualized and evaluated. They note that current evaluation practices primarily probe LLMs for ``Cultural awareness'', i.e., their culture-specific knowledge and reasoning capabilities, by relying on curated cultural test beds. However, they argue, performing well on such benchmarks solely demonstrates the knowledge of the cultures that are tested for and does not demonstrate the ability to operate in previously unseen cultural contexts.
 Instead, they propose the concept of \textit{meta-cultural competence}, which refers to an AI system’s ability to recognize cultural variation and adapt to new cultural contexts. While this perspective clarifies the long-term capability that culturally robust AI systems should aspire to, it leaves open the question of what levels of cultural capabilities should be defined and measured in current NLP evaluations. The goal of our work is complementary to that of \citet{saha-etal-2025-meta}. Rather than proposing a new target capability, we focus on defining different levels of cultural capability, drawing on intercultural communication research, to improve construct clarity and measurement validity in cultural evaluation.

We echo the observation by \citet{saha-etal-2025-meta} that most benchmarks concerned with cultural inclusivity are focused on measuring ``knowledge about a cultural context''. Examples include \textsc{FORK} \citep{palta-rudinger-2023-fork}, which targets food-related cultural commonsense such as ingredients, preparation methods, and culturally appropriate consumption practices; \textsc{CulturalBench} \citep{chiu-etal-2025-culturalbench}, which introduces region-specific multiple-choice questions covering everyday activities, social norms, public behavior, and local conventions; and \textsc{BLEnD} \citep{myung-etal-2024-blend}, which focuses on everyday practices and social routines (e.g., food, sports, family, holidays/celebrations/leisure) across 16 regions and 13 languages. \textsc{GeoMLAMA} \citep{yin-etal-2022-geomlama} probes geo-diverse commonsense knowledge,  concepts that are universally understood but vary across different cultures and regions, such as the color of a traditional wedding dress, staple foods and units of measurement. \textsc{INCLUDE} \citep{romanou2025include}, on the other hand,  curates exam-style questions in 44 languages that emphasize culturally situated general knowledge and reasoning skills.  \textsc{JMMMU} \citep{onohara-etal-2025-jmmmu} is another work in this line, which incorporates multimodal cultural knowledge in domains such as arts and heritage.

Several recent works attempt to operationalize cultural understanding as recognition of culturally inappropriate signals. One example is \textsc{MC-SIGNS} by \citet{yerukola-etal-2025-mind}, which evaluates whether models can classify gestures as offensive or non-offensive depending on the cultural context. Other resources foreground stereotypical statements about social groups, such as \textsc{SHADES} \citep{mitchell-etal-2025-shades}, which evaluates stereotypes across regions and languages, spanning multiple identity categories subject to discrimination. \citet{qiu-etal-2025-evaluating} evaluate agents' ability to detect and appropriately respond to norm-violating user queries and observations, for online shopping and social discussion forums. 

More recent work attempts to evaluate cultural capabilities in interactive settings. \textsc{NormGenesis} \citep{hong-etal-2025-normgenesis} goes beyond knowledge by measuring culturally adaptive dialogue in multi-turn conversations, focusing on the integration of social norms into interactional behavior. \textsc{Nunchi-Bench} \cite{kim-lee-2025-nunchi} is another benchmark containing scenario-based questions that require models to identify culturally appropriate responses or explanations. \textsc{SocialCC} by \citet{wu-etal-2025-socialcc} evaluates LLM performance in multi-turn social interactions where appropriate responses depend on cultural norms and contextual cues, and measures whether models produce socially appropriate responses. Similarly, \citet{havaldar-etal-2025-culturally} propose a framework for evaluating the cultural awareness of language models in multicultural conversational environments. Their evaluation incorporates situational context, interpersonal relationships, and conversational style to assess how well models adapt to culturally grounded interactions. These works represent an important step toward evaluating cultural competence as a dynamic capability rather than static knowledge.

\noindent \textbf{Gap Analysis:}
Although the discussion above does not constitute a systematic literature review of cultural capability evaluations in NLP, it nevertheless reveals substantial evidence of construct ambiguity in the current literature. Across these works, terminology referring to cultural capability dimensions is highly inconsistent and often underspecified. Terms such as ``cultural understanding,'' ``cultural adaptation,'' ``cultural awareness,'' ``cultural sensitivity,'' and ``cultural competence'' are frequently used interchangeably, sometimes even within the same work, without precise definitions or explicit alignment with established social science theories. As a result, different studies implicitly measure different aspects of cultural behavior while referring to them using fuzzy terminology. Because of this fundamental lack of construct validity, it becomes unclear what capability an evaluation actually measures and whether results across benchmarks are comparable. Consequently, evaluation results are often interpreted as evidence of ``cultural capability'' in general, even though they may only capture a narrow dimension of that construct.

What is therefore needed is a framework that explicitly distinguishes between different levels of cultural capability and provides clear definitions of what each level entails in terms of observable system behavior. Such a framework would enable researchers to select the level of capability relevant to their task, design evaluation procedures that directly measure that capability, and make appropriately scoped claims about model performance.

\section{Evaluative Models of Cultural Capabilities in ICC}
\label{sec:icc-models}

Intercultural communication research has long emphasized that effective engagement across cultures requires more than static knowledge of norms or practices. Across several influential models, scholars have conceptualized ``cultural capabilities'' as multidimensional constructs encompassing cognitive, affective, and behavioral components. We review three foundational and highly cited ICC traditions: the Developmental Model of Intercultural Sensitivity (DMIS), the theory of Cultural Intelligence (CQ), and the Process Model of Intercultural Competence (PMIC). For each ICC model, we discuss 1) a focal capability, 2) a structure for that capability (whether stages, dimensions, or component skills), and 3) sites of application with corresponding measurement strategies. Table~\ref{tab:eval_models} summarizes the characteristics of these models.

\begin{table*}[ht]
\small
\centering
\renewcommand{\arraystretch}{1.35}
\setlength{\tabcolsep}{6pt}

\begin{tabular}{p{2.3cm} p{4cm} p{3.5cm} p{4cm}}
\toprule
\textbf{Model} & \textbf{Focal Cultural Capability} & \textbf{Structure} & \textbf{Evaluation} \\
\midrule

\textbf{DMIS} \newline \citep{bennett1986developmental, bennett1993towards} &
\textbf{Sensitivity:} How individuals experience and interpret cultural differences. &
Six stages from ethnocentrism (\textit{Denial, Defence, Minimization}) to ethnorelativism (\textit{Acceptance, Adaptation, Integration}). &
\textit{Intercultural Development Inventory (IDI)}. \\

\textbf{CQ} \newline \citep{earley2003cultural, ang2007cultural} &
\textbf{Intelligence:} Capability to function effectively across diverse cultural contexts. &
Four dimensions: \textit{Motivational, Cognitive, Metacognitive, Behavioral}. &
\textit{Cultural Intelligence Scale (CQS)}. \\

\textbf{PMIC} \newline \citep{deardorff2006identification, deardorff2009synthesizing} &
\textbf{Competence:} Ability to communicate effectively and appropriately across cultures. &
Cyclical model linking \textit{Attitudes}, \textit{Knowledge}, \textit{Skills}, producing \textit{Internal/External Outcomes}. &
\textit{ICA} and AAC\&U \textit{VALUE Rubric}. \\
\bottomrule
\end{tabular}

\caption{Summary of three major ICC models frequently used for evaluating cultural capabilities.}
\label{tab:eval_models}
\end{table*}

\subsection{Developmental Model of Intercultural Sensitivity (DMIS)}

\vspace{5pt}

\noindent \textbf{Focal Capability:} DMIS \cite{bennett1986developmental} is one of the earliest evaluative ICC models and is focused on \textit{intercultural sensitivity} as the core capability, which refers to the way individuals \textit{experience} and \textit{make sense of} cultural differences. This model is also inherently developmental, i.e., it proposes that individuals progress through qualitatively different stages of worldview, moving from ethnocentrism toward ethnorelativism \citep{bennett1993towards}.  

\vspace{5pt}

\noindent \textbf{Structure: } DMIS describes \textit{intercultural sensitivity} as a sequence of stages. The ethnocentric stages include \textit{1) Denial} (lack of recognition of cultural difference), \textit{2) Defence} (perceiving difference as threatening and asserting superiority of one's own culture), and \textit{3) Minimization} (downplaying difference by assuming deep similarity or universalism). As intercultural sensitivity increases, people move towards the ethnorelative stages, namely, \textit{4) Acceptance} (recognition and valuing of cultural difference), \textit{5) Adaptation} (the ability to shift perspective and modify behavior appropriately), and \textit{6) Integration} (internalization of multiple cultural perspectives into one's own identity). 

\vspace{5pt}

\noindent \textbf{Application and Evaluation: } DMIS is applied in international education, study abroad, and professional development for people working in multicultural contexts, such as health care providers \cite{pedersen2010assessing, dejaeghere2009developing, bourjolly2005journey, richards2016intercultural}. Measurement is often done using the Intercultural Development Inventory (IDI), which attempts to position individuals along a continuum from \textit{Denial} to \textit{Integration} through survey items targeting beliefs, reactions, and self-perceived adaptability.

\subsection{Cultural Intelligence (CQ)}


\vspace{5pt}

\noindent \textbf{Focal Capability:} The CQ model \cite{earley2003cultural} emerged to reduce costly failures in international assignments caused by stereotyping and cultural generalizations \citep{black1991toward,mendenhall2008specification} and defines \textit{cultural intelligence} as an individual's capability to function effectively in situations characterized by cultural diversity. 

\vspace{5pt}

\noindent \textbf{Structure:} CQ is explicitly framed as a \textit{multidimensional intelligence} and distinguishes four interrelated capabilities: \textit{1) Motivation} (drive to engage across cultures), \textit{2) Cognition }(knowledge of cultural norms, practices), \textit{3) Metacognition} (awareness of and ability to plan, monitor, and adjust one’s thought processes in intercultural interactions), and \textit{4) Behavior} (ability to adapt one’s verbal/nonverbal conduct such as adapting tone, turn-taking patterns, politeness strategies, gesture, pace, etc.) in culturally diverse interactions \citep{ang2007cultural,ang2015handbook}. 

\vspace{5pt}

\noindent \textbf{Application and Evaluation:} 
\enlargethispage{1\baselineskip}
CQ is applied in leadership development, international assignments, 
and cross-border negotiation \citep{alon2005global,rockstuhl2011beyond,ramalu2012cultural}. Higher CQ is associated with 
better task performance in culturally diverse settings \citep{ang2007cultural} and is linked to experiential learning theory \citep{kolb2014experiential}. 
 CQ is typically measured through validated psychometric instruments such as the Cultural Intelligence Scale (CQS), which measures each dimension on a Likert scale and has been adapted and validated cross-nationally \cite{van2015development,gozzoli2018cultural}. 

\subsection{Process Model of Intercultural Competence (PMIC)}


\vspace{5pt}

\noindent \textbf{Focal Capability:} PMIC \cite{deardorff2006identification} conceptualizes intercultural competence as a dynamic, iterative process and defines \textit{intercultural competence} as ``the ability to communicate effectively and appropriately in intercultural situations based on one’s intercultural knowledge, skills, and attitudes''. This view integrates both developmental and performance-based perspectives and recognizes that competence manifests in interaction rather than merely in perception or cognition.

\vspace{5pt}

\noindent \textbf{Structure:} 
PMIC proposes a cyclical relationship among five interrelated components:  \textit{1) Attitudes} (respect, openness, curiosity, willingness to tolerate ambiguity);  \textit{2) Knowledge} (including self-awareness, deep cultural knowledge, and sociolinguistic awareness); \textit{3) Skills} (listening, observing, analyzing, evaluating, and relating); \textit{4) Internal Outcomes} (adaptability, flexibility, empathy, ethnorelative view) leading to \textit{5) External Outcomes} (effective and appropriate behavior and communication). Importantly, \citet{deardorff2009sage} emphasizes that the process is ongoing, recursive, and context-dependent, allowing for continuous development through experience and reflection.

\vspace{5pt}

\noindent \textbf{Applications and Evaluation:} 
PMIC is extensively applied in higher education, internationalization of curricula, global citizenship education, and intercultural training across disciplines such as health, business, and diplomacy \citep{byram2020teaching, arasaratnam2017intercultural}.
Building on her process model, \citet{deardorff2006identification} developed the \textit{Intercultural Competence Assessment (ICA)} framework and later contributed to the \textit{Intercultural Knowledge and Competence VALUE Rubric} \cite{AACU_InquiryAnalysis_VALUE_Rubric}. These tools are primarily qualitative and reflective rather than psychometric 
\citep{deardorff2009synthesizing}.

\section{A Taxonomy of AI-Relevant Cultural Capabilities}
\label{sec:taxonomy}

Here, we propose a taxonomy of \emph{required} and \emph{measurable} cultural capabilities in AI-enabled communication and ground this taxonomy in ICC models described in Section~\ref{sec:icc-models}. For that, we first recognize that the three major evaluative ICC models were developed to describe \emph{human} experience, motivation, and behavior, and the direct application of these models to AI systems risks anthropomorphizing. Therefore, we deliberately choose a cautious starting point and treat these models as \emph{conceptual resources} rather than as templates to be copied. As a result of this choice, in our work, \textbf{capability} refers to observable behavior that is elicited in a particular interaction, as opposed to a trait that the model has independent of the interaction context.

Following literature that shows large language models do not possess a stable moral or normative stance \cite{abdulhai2024moral,guo2024adaptable}, we restrict our taxonomy to traits that are observable in the \emph{linguistic behavior} of AI systems. While human-focused models of cultural competence consider ``worldviews'', ``attitudes'', or ``motivation'', we do not assume that AI shares any analogous internal orientation. Instead, to avoid overclaiming about AI's cultural capabilities, we ask a narrower question: \emph{which aspects of these constructs have recognizable linguistic footprints that can appear in model outputs and be evaluated as such?}

Concretely, we reinterpret the constructs in DMIS, CQ, and PMIC as a mixture of (a) \emph{motivational} components, which are intrinsically tied to human agency and affect, and (b) \emph{behavioral} components, which manifest in discourse, framing, and interactional patterns. While both classes matter for humans, for AI, only the latter can be meaningfully operationalized. 

Our methodology is divided into three steps. In Step 1, we identify, within each model, which elements have observable linguistic manifestations. In Step 2, we recategorize the observable behaviors into distinct levels of capabilities. In Step 3, we re-interpret these levels of capability for AI.

\vspace{5pt}

\noindent \textbf{Step 1:} In the following, across the ICC models, we distinguish between \emph{motivational} (human-only) and \emph{behavioral} elements (human and AI):

\vspace{5pt}

 \textbf{DMIS:} Although DMIS stages are originally framed as developmental worldviews, we argue that these stages also have recognizable \emph{discursive correlates}. For example, \textit{Denial} can surface as linguistic erasure of difference (\textit{``people everywhere are basically the same''}), \textit{Defence} as superiority framing (\textit{``our way is more advanced''}), and \textit{Minimization} as universalizing language (\textit{``deep down, all cultures want the same things''}). \textit{Acceptance} and \textit{Integration} manifest in explicit acknowledgments of difference and multi-perspective framing, while \textit{Adaptation} involves shifts in tone, register, or politeness strategies. We therefore treat DMIS stages as \emph{behavioral} elements for AI, even though AI does not inherently possess those worldviews.

\vspace{5pt}

\textbf{CQ:} We categorize the \textit{Motivational} element of CQ as a human-only construct that is inherently tied to human intention and effort. By contrast, \textit{Cognitive} CQ (knowledge of norms and practices) can appear in model outputs as factual recall and distinctions between cultural practices. \textit{Metacognitive} CQ (planning, monitoring, and adjusting one's interpretation) has also partial behavioral manifestations in AI when models provide reasoning, reconsider earlier assumptions, or explicitly hedge and revise interpretations. Finally, \textit{behavioral} CQ, the ability to adapt verbal behavior across contexts, can be observed in text as shifts in tone, politeness, register, or interactional style. These three CQ components thus contribute directly to AI-relevant behavioral capabilities.

\vspace{5pt}

 \textbf{PMIC:} We argue that the elements of \textit{Attitudes} and \textit{Internal Outcomes} in PMIC are explicitly affective and experiential; we again treat them as human-only traits and avoid projecting them onto AI systems. By contrast, \textit{Knowledge} (cultural knowledge and sociolinguistic awareness), together with \textit{Skills} (observing, analyzing, relating, evaluating), can be observed in discourse as the ability to describe, interpret, and compare cultural practices. Lastly, \textit{External Outcomes} correspond to effective and appropriate behavior and communication in intercultural encounters, which can be evaluated for AI systems via their response content, tone, and pragmatic appropriateness.

\vspace{5pt}

\noindent \textbf{Step 2:} We restrict attention to observable behaviors based on the above analysis and recategorize them to obtain a single taxonomy. Across DMIS, CQ, and PMIC, intercultural effectiveness is consistently decomposed into three broad families of observable\textit{ human} capabilities, which we describe first below and reinterpret in Step 3 for AI.

\vspace{5pt}

 \textbf{Cognitive foundations:} the informational substrate of intercultural behavior, including knowledge, awareness, and understanding of cultural differences (cognitive CQ; Knowledge in PMIC), such as accurate descriptions of practices, recognition of group-specific norms, and sociolinguistic knowledge (e.g., honorifics, forms of address). 

 \vspace{5pt}

\textbf{Framing and stance-taking:} the ways in which cultural differences are \emph{positioned} and \emph{expressed} in discourse. This draws on DMIS stages as observable stances (\textit{Denial, Defence, Minimization, Acceptance, Integration})\footnote{We omit \textit{Adaptation} here because it is captured under interactional adaptation later.} and on PMIC's emphasis on appropriateness. 

\vspace{5pt}

 \textbf{Interactional adaptation:} the competence and skills required to adjust communication in situ, across turns and evolving contexts. This includes \textit{behavioral} CQ and \textit{Metacognitive} CQ as well as \textit{Skills} and \textit{External Outcomes} of PMIC. These skills can manifest as shifting tone, register, or explanatory strategy when new cultural cues emerge; revising an explanation when the user signals discomfort; and coordinating meaning over time rather than in a single shot. 

\vspace{5pt}

\noindent \textbf{Step 3:} Building on this behavioral reinterpretation, we articulate three AI capability levels that align with, but do not collapse into, the behavioral human-focused constructs, and are empirically testable with NLP methods (Figure \ref{fig:taxonomy}).

\vspace{5pt}

 \textbf{Capability Level 1: Cultural Awareness - } This level concerns the model's ability to represent and retrieve culture-specific information accurately. It corresponds primarily to the cognitive foundations drawn from CQ and PMIC: factual knowledge about practices, norms, histories, and sociolinguistic conventions. Evaluations at this level target informational accuracy and coverage: does the model correctly distinguish between different cultural practices, avoid hallucinating non-existent customs, and resist collapsing distinct groups into monolithic categories?

\vspace{5pt}

 \textbf{Capability Level 2: Cultural Sensitivity -} This level concerns the model's ability to frame cultural differences respectfully and non-ethnocentrically. It is a one-shot property of the model's initial stance toward cultural cues in the prompt and is grounded in the behavioral readings of DMIS stages and PMIC's focus on appropriateness. Here, the question is not yet whether the model can adapt over time, but whether its first move avoids \textit{Denial, Defence, or Minimization} and instead recognizes difference without othering. Evaluations at this level focus on stance and framing: whose perspective is centered, what is normalized, and whether the language implicitly ranks cultures.

    \vspace{5pt}

 \textbf{Capability Level 3: Cultural Competence -} This level concerns the model's ability to adapt its communicative behavior dynamically as the interaction unfolds and new cultural cues emerge. It includes interactional adaptation capabilities: perspective-shifting, pragmatic adjustment, and context-sensitive revisions across multiple turns. A culturally competent model should not only begin from a non-harmful stance but also update its responses when a user signals a particular identity, constraint, or harm history. Evaluations at this level require multi-turn setups and focus on dynamic behavior: how responses evolve, whether the model corrects earlier misframings, and how it coordinates meaning with the user over time.

\section{Application of Taxonomy in AI Evaluation}



While various dimensions of cultural capabilities have been measured by AI researchers, the terminologies used to describe these dimensions are often underspecified and used interchangeably. Our taxonomy provides an ICC-grounded vocabulary that enables researchers to identify and describe the level of cultural capability being measured in a more systematic way. This taxonomy is a practical tool for evaluators of AI systems to 1) specify which cultural capabilities a given task requires before designing the evaluation, 2) design evaluations that target the corresponding observable behaviors, and 3) clarify what the evaluations do not capture. For example, for a narrowly focused question-answering system, \textit{diverse factual knowledge} is the minimum required level of cultural capability; the evaluations need to capture a wide coverage of culturally-grounded QA tests. Scoring high on such tests demonstrates \textit{Cultural Awareness}, but the model might still lack \textit{Cultural Sensitivity} (might use ethnocentric framing) or \textit{Cultural Competence} (fail to adapt when the context changes). 
When the level of cultural capability being measured is not explicitly specified, these results may be misinterpreted and mislead the decision makers. 


In some tasks, all levels of cultural capabilities are required. For a real-world example, consider a conversational system used in K–12 education (for instance, see \citet{unesco2025whatsapp} for developing such a chatbot in Zimbabwe). Such a system is required to demonstrate all three levels of cultural capabilities identified in our taxonomy.  Consider the query \emph{``Why do some communities prefer spiritual healing methods over clinical treatments?''}. A \textit{Culturally Aware} model accurately describes practices, contexts, and underlying cultural reasoning, avoiding factual errors. A \textit{Culturally Sensitive} model frames cultural differences with respect, avoids ethnocentric or moralizing language, and explicitly recognizes cultural specificity while remaining educational and informative. After the initial answer, the user clarifies: \emph{``In my community, we rely heavily on herbal remedies and rituals, and some people worry that modern medicine dismisses them.''} A \textit{Culturally Competent} model adjusts tone and framing to reflect the user’s perspective, mediates between potentially conflicting epistemologies, recovers from initial assumptions, and maintains consistent respect and accuracy across multiple turns. Therefore, the evaluation of this system needs to tackle all these criteria at all three levels. 
 
Once the required level of capability is identified, researchers need to align evaluation designs with the required capability levels. To evaluate \textit{Awareness}, culturally grounded knowledge benchmarks, stereotype audits, and multi-regional and multi-lingual QA tests are sufficient. Representative examples of NLP work that measures \textit{Awareness}, as defined in our taxonomy, include \textsc{GeoMLAMA} \cite{yin-etal-2022-geomlama}, \textsc{FORK} \cite{palta-rudinger-2023-fork}, \textsc{BLEnD} \cite{myung-etal-2024-blend}, \textsc{INCLUDE} \cite{romanou2025include}, and  \textsc{CulturalBench} \cite{chiu-etal-2025-culturalbench}. Evaluating \textit{ Sensitivity} is facilitated through single-turn prompts annotated for tone, stance, and framing by intercultural experts; probes that inspect how the model describes or contrasts cultural differences. Relevant resources include \textsc{SHADES} \cite{mitchell-etal-2025-shades}, which measures stereotype framing across languages, and \textsc{MC-SIGNS} \cite{yerukola-etal-2025-mind}, which was developed to detect culturally offensive signals.

Arguably, evaluating \textit{Competence} is more challenging than the other two levels and can only be achieved through multi-turn simulations and user-in-the-loop studies that assess whether the model adjusts to new cultural cues, resolves ambiguity, and repairs misalignment over time. Such evaluations can be operationalized as scenario-based dialogues in which a culturally salient cue is introduced after the model's initial response. For example, the user discloses their community, a religious constraint or a local practice, and the model is scored on whether the subsequent turns revise prior assumptions, produce necessary clarification, or accommodate the new information in another way. Examples of NLP works that do evaluate competence, as defined in our paper (although they might use other terms to refer to it), are as follows: \textsc{NormGenesis} \cite{hong-etal-2025-normgenesis} offers one template by tracking the integration of social norms across turns; \textsc{SocialCC} \cite{wu-etal-2025-socialcc} and the framework by \citet{havaldar-etal-2025-culturally} extend this to socially situated multi-turn exchanges; and \textsc{Nunchi-Bench} \cite{kim-lee-2025-nunchi} provides scenario-based prompts that could be extended into multi-turn variants. Appropriate systematized metrics should be developed to measure desired behaviors such as whether the model explicitly references the user-introduced cultural cue in subsequent turns, whether earlier ethnocentric or generic framings are repaired without further prompting, or whether respectful framing is maintained as the conversation evolves. Designing such evaluations for low-resource languages will require participatory methods and community partnerships, since model behavior in these settings is constrained by training-data coverage.

Future work should focus on developing NLP methods capable of detecting the signals associated with each level of cultural capability within a given interaction. For example, the rich bodies of work on bias detection \cite{field2021survey}, counter-stereotype generation \cite{zheng2023makes,fraser2023makes,nejadgholi2024challenging}, stance detection \cite{kuccuk2020stance}, and affective computing \cite{pei2024affective} provide methodological foundations for operationalizing the more complex levels of cultural capability, particularly adaptive cultural competence, which requires models to interpret users’ evolving cues, adjust tone, and modulate responses dynamically.

\section{Conclusion}


To address construct ambiguity in evaluating AI’s cultural capabilities, we introduce a taxonomy grounded in intercultural communication theory that distinguishes between Cultural Awareness, Sensitivity, and Competence, and frames them in terms of observable system behavior.

We argue that improving construct clarity is essential for reliable evaluation in practice. When cultural capability is underspecified, evaluation results may overestimate model readiness, particularly when knowledge-based performance is interpreted as broader competence. 
We therefore encourage more explicit, capability-aligned evaluation practices that clarify what is being measured and what is not, particularly in multicultural contexts where the consequences of misinterpretation are amplified.
\section*{Limitations}

 It is important to note that rigorous measurement alone cannot resolve the broader sociotechnical harms associated with English-centric AI-mediated communication. As \citet{wallach2024evaluating} caution, even well-structured measurement frameworks do not automatically translate into better outcomes; rather, they make explicit what evaluations capture and, equally importantly, what they omit. We adopt this perspective in our work, using conceptual systematization as a means to clarify which aspects of cultural capability are being measured in AI evaluation and which remain outside the scope of measurement.

Additionally, the taxonomy proposed in this work should not be interpreted as a comprehensive account of all cultural capabilities relevant to AI systems. Intercultural communication is a complex and multidimensional phenomenon studied across several disciplines, including communication studies, sociology, education, and social psychology. As such, additional constructs and distinctions may emerge as research on culturally grounded AI evaluation evolves. Therefore, we did not exhaustively enumerate all possible cultural capabilities, but addressed a specific gap in the current NLP literature: the conceptual ambiguity surrounding the terminology used to describe cultural capabilities. 

Another limitation arises from the ICC models, on which we base our taxonomy. DMIS, CQ, and PMIC were developed primarily in workplace, education, and expatriate-adjustment contexts, and as a result emphasize an ``outsider'' view of culture. Real-world users of AI, however, might seek support in navigating their own social relationships, from an ``insider'' view of culture. Extending our taxonomy toward insider-oriented competence would depend on participatory and community-informed methods, narrative-based scenarios, and evaluators with lived cultural experiences.

Further, given the fluid and evolving nature of both ``culture'' and ``cultural groups'', complete knowledge of norms and variations associated with all cultural boundaries might be an impossible goal. An important cognitive ability defined in the ICC literature is \textit{metacognition}: identifying situation-relevant norms that may be culture-specific and obtaining missing information before formulating a final response, rather than assuming a universal norm. This higher level of metacognitive behaviors in intercultural interactions, where one shifts from assuming normative cultural standards to recognizing and adapting behaviors based on incoming conversational cues, is challenging and is currently understudied in the landscape of cross-cultural AI evaluations.

Finally, the boundaries between the levels in our taxonomy, Awareness, Sensitivity, and Competence, should not be interpreted as rigid or mutually exclusive categories. In practice, these capabilities often interact and may appear simultaneously in system behavior. The taxonomy is therefore best understood as a conceptual scaffold that helps researchers articulate which aspect of cultural capability an evaluation targets, rather than as a definitive or exhaustive model. Future work may refine, expand, or reorganize these categories as empirical evidence and interdisciplinary insights accumulate.

\bibliography{references}

\appendix

\end{document}